\documentclass[10pt,twocolumn,letterpaper]{article}

\usepackage[pagenumbers]{cvpr} 

\usepackage{graphicx}
\usepackage{amsmath}
\usepackage{amssymb}
\usepackage{booktabs}
\usepackage{algorithm}
\usepackage{algorithmic}
\usepackage{caption}
\usepackage{subcaption}

\usepackage[pagebackref,breaklinks,colorlinks]{hyperref}

\usepackage[capitalize]{cleveref}
\crefname{section}{Sec.}{Secs.}
\Crefname{section}{Section}{Sections}
\Crefname{table}{Table}{Tables}
\crefname{table}{Tab.}{Tabs.}

\def\cvprPaperID{12057} 
\def\confName{CVPR}
\def\confYear{2022}

\begin{document}

\title{Accelerating Deep Learning with Dynamic Data Pruning}

\author{Ravi Raju\\
University of Wisconsin-Madison\\
1405 Engineering Dr, Madison, WI, 53706 \\
{\tt\small rraju2@wisc.edu}
\and
Kyle Daruwalla\\
University of Wisconsin-Madison\\
1405 Engineering Dr, Madison, WI, 53706 \\
{\tt\small daruwalla@wisc.edu}
\and
Mikko Lipasti\\
University of Wisconsin-Madison\\
1405 Engineering Dr, Madison, WI, 53706 \\
{\tt\small mikko@engr.wisc.edu}
}

\maketitle

\begin{abstract}
Deep learning’s success has been attributed to the training of large, overparameterized models on massive amounts of data. As this trend continues, model training has become prohibitively costly, requiring access to powerful computing systems to train state-of-the-art networks. A large body of research has been devoted to addressing the cost per iteration of training through various model compression techniques like pruning and quantization. Less effort has been spent targeting the number of iterations. Previous work, such as forget scores and GraNd/EL2N scores, address this problem by identifying important samples within a full dataset and pruning the remaining samples, thereby reducing the iterations per epoch. Though these methods decrease the training time, they use expensive static scoring algorithms prior to training. When accounting for the scoring mechanism, the total run time is often increased. In this work, we address this shortcoming with dynamic data pruning algorithms. Surprisingly, we find that uniform random dynamic pruning can outperform the prior work at aggressive pruning rates. We attribute this to the existence of ``sometimes'' samples --- points that are important to the learned decision boundary only some of the training time. To better exploit the subtlety of sometimes samples, we propose two algorithms, based on reinforcement learning techniques, to dynamically prune samples and achieve even higher accuracy than the random dynamic method. We test all our methods against a full-dataset baseline and the prior work on CIFAR-10 and CIFAR-100, and we can reduce the training time by up to 2$\times$ without significant performance loss. Our results suggest that data pruning should be understood as a dynamic process that is closely tied to a model's training trajectory, instead of a static step based solely on the dataset alone.
\end{abstract}
\section{Introduction}
\label{sec:introduction}

A growing body of literature recognizes the immense scale of modern deep learning (DL) \cite{bommasani_opportunities_2021, strubell_energy_2019}, both in model complexity and dataset size. The DL training paradigm utilizes clusters of GPUs and special accelerators for days or weeks at a time. This trend deters independent researchers from applying state-of-the-art techniques to novel datasets and applications, and even large research organizations accept this approach at significant costs.

\begin{figure}
    \centering
    \includegraphics[width=0.45\textwidth]{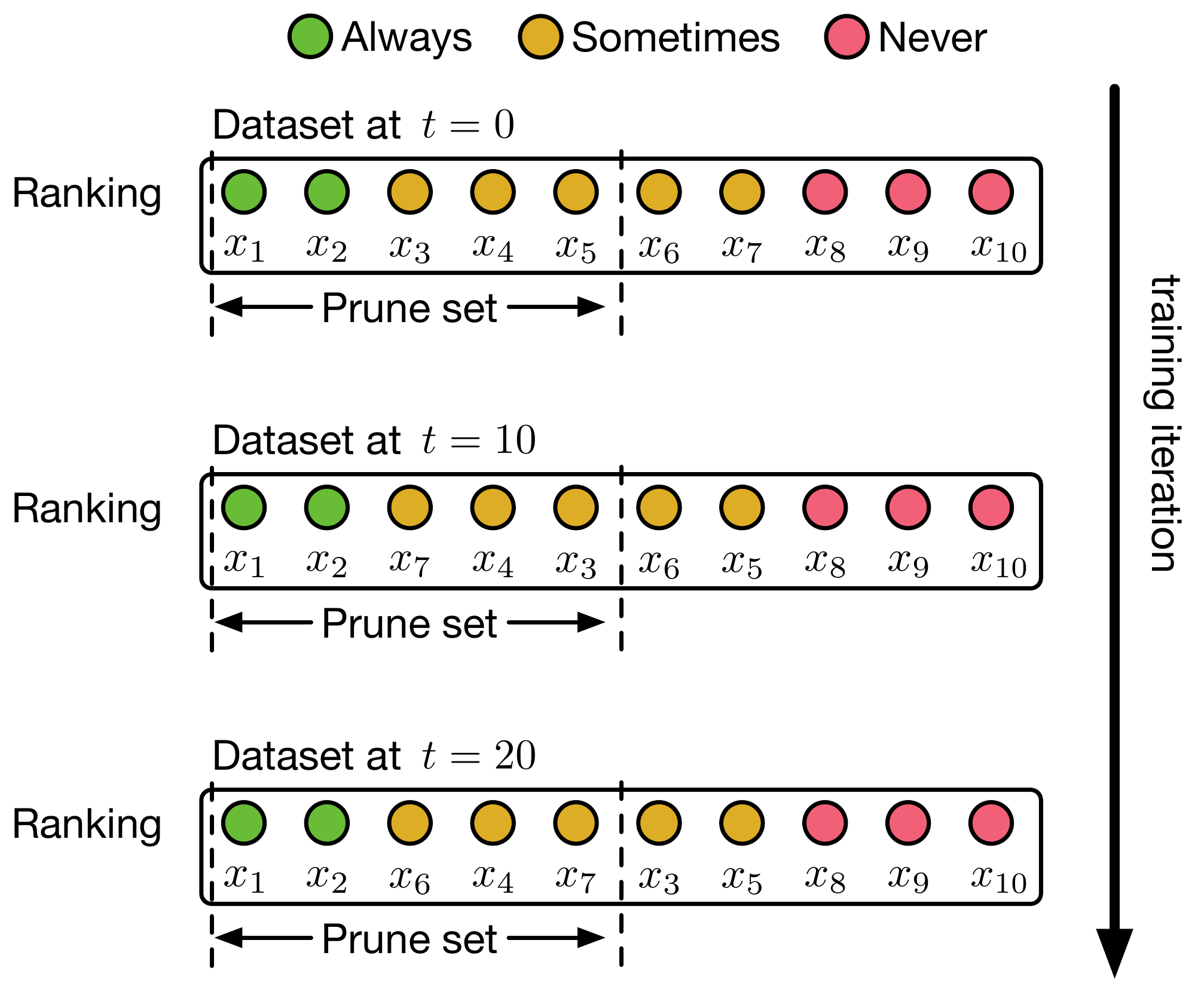}
    \caption{Dynamic data pruning presents multiple opportunities for a sample to be selected for training. This illustration shows our finding that datasets are separated into three groups --- samples that are always selected, samples that are never selected, and samples that are selected only some of the time. Static methods fail to effectively target the last group, since their ranking varies with training.}
    \label{fig:sometimes-set}
\end{figure}

Currently accepted methods \cite{narang_mixed_2018} target the per-iteration penalty of evaluating the model during training; however, not as much effort has been on spent reducing the total number of training iterations. Since even simple datasets \cite{krizhevsky_learning_2009} require hundreds of epochs over tens of thousands of samples, eliminating a non-essential subset of data presents a promising opportunity for efficiency. 

Work from other DL domains suggest that only subset of the data influences the decision boundary and contributes a majority of the loss incurred by the model \cite{agarwal_estimating_2021, toneva_empirical_2019}. Furthermore, curriculum learning \cite{bengio_curriculum_2009} asserts that samples can be ranked which might allow us to prune redundant ``easy'' samples. Prior work on data pruning \cite{toneva_empirical_2019, paul_deep_2021} take advantage of this property to eliminate a majority of the dataset without incurring significant performance loss. Unfortunately, these methods run their scoring algorithm prior to training and require one or more passes over the dataset. When we include the cost of scoring the samples, the total run time exceeds the time it takes to do a single conventional training run. This prevents researchers from utilizing the prior work on new, non-standard datasets.

In our work, we dynamically select a subset of the data at fixed checkpoints throughout training. We make the following new observations and contributions:
\begin{enumerate}
    \item By counting every time each sample is selected across all scoring checkpoints, we find that a dataset can be qualitatively split into three groups --- always samples, never samples, and sometimes samples (see \cref{fig:sometimes-set}). Always samples are selected at nearly every scoring checkpoint. Similarly, never samples are rarely selected. Sometimes samples are selected at only some checkpoints, and their inclusion is highly variable across different training runs. Static pruning methods, like the prior work, can identify always or never samples but fail to effectively target sometimes samples. In fact, we find that randomly selecting a subset of the data at each checkpoint is more effective than the static baselines.
    \item Given this grouping of the dataset, we design a dynamic scoring mechanism based on per-sample loss. Despite scoring more frequently, our mechanism reduces the total run time including the cost of scoring, while the prior work typically increases it. Moreover, at aggressive pruning rates, we obtain higher final test accuracies on CIFAR-10, CIFAR-100, and a synthetically imbalanced variant of CIFAR-10.
    \item Since the sometimes samples vary in importance across training runs, we note that the optimal dynamic scoring selection is tightly coupled to the model trajectory. So, we re-frame the data pruning problem as a decision making process. Under this lens, we propose two variants of our scoring mechanism that borrow from $\epsilon$-greedy and upper confidence bound (UCB) algorithms in reinforcement learning. With these additional improvements, we obtain even higher performance at aggressive pruning rates even when the dataset is imbalanced.
\end{enumerate}
\section{Related Work}
\subsection{Curriculum learning and active learning}
The idea that certain samples are more influential than others has been observed for some time in curriculum learning and active learning. Curriculum learning exploits this to present samples to the model under a favorable schedule during training to obtain better generalization \cite{bengio_curriculum_2009, cai_curriculum_2018}. Our work has a similar objective, but we are concerned with ranking samples in order to eliminate them instead of schedule them.

Active learning attempts to rank new, un-labeled samples based on redundancy instead of difficulty. One of the computationally cheap techniques in this setting is uncertainty sampling \cite{settles_active_2009}, which we use in our work as well. Unlike active learning, in data pruning we have access to all the samples and their labels.


\subsection{Static data pruning}
Several works have studied how to statically prune a dataset prior to training. Toneva \etal \cite{toneva_empirical_2019} proposed the concept of forget scores, which track the number of times an example transitions from classified correctly to misclassified throughout training. They classified the data into two sets --- forgettable examples (frequently misclassified) and unforgettable examples (rarely misclassified). By pruning samples based on the forget scores, they eliminate up to 30\% of the data with no loss in performance on CIFAR-10. An important limitation of their method is the need to perform a full training run to obtain the forget scores. Coleman \etal \cite{coleman_selection_2020} build on forget scores in their work called selection via proxy (SVP). SVP uses a smaller proxy model to rank examples in order to choose a subset that can be used to train a larger model. By re-using the forget scores algorithm and training a smaller network for fewer epochs, they obtain a subset of samples to train a larger model with no loss in performance even when the subset is 50\% of the original dataset size. Unlike the forget score-based methods, Paul \etal \cite{paul_deep_2021} use the gradient norm-based (GraNd) scores of individual samples averaged across 10 different random models to prune data at initialization. They also prune up to 50\% of CIFAR-10 with no loss in accuracy. Even though the SVP and GraNd methods require less computation than the original forget score work to obtain a ranking, the results are too noisy to statically prune the data. Thus, they must rely on more iterations or averaging across trials to obtain the final subset of samples. As a result, all three methods incur a high overhead for pruning that increases the total run time to train the final model. This prevents these methods from being applied to new datasets for the first time. Since our method is dynamic, it can be applied directly to new data without \emph{a priori} information.
\section{Methods}

In this section, we describe our approach to data pruning. First, we re-frame data pruning as a dynamic decision making process. Next, we present our scoring mechanism based on filtered uncertainty sampling. Finally, we discuss how the $\epsilon$-greedy and upper confidence bound algorithms used in reinforcement learning can be used on top of our scoring mechanism to further improve the final test accuracy.

\subsection{Dynamic data pruning}

Static data pruning assumes that the optimal subset of samples depends primarily on the data distribution. Model training is a means to an end, and the trajectory of a specific model does not factor into the final pruned dataset. In our work, instead of approaching data pruning as a one-time step, we view it as a dynamic process that is coupled to a given model's training run. The total number of training epochs is divided by fixed pruning checkpoints. At each checkpoint, an algorithm may observe the current performance of the model on the full dataset. Given this observation and information retained from previous checkpoints, the algorithm makes a decision on which points to use for training until the next checkpoint. This process is described in \cref{alg:dynamic-pruning}.

\begin{algorithm}
\caption{Dynamic data pruning process}
\label{alg:dynamic-pruning}
\begin{algorithmic}
\REQUIRE $k$ = \# of samples to keep, $s$ = pruning selection criterion, $T_p$ = pruning period, $T$ = total \# of epochs, dataset $X$ (w/ $N$ samples), classifier $f_{\theta}$
\FOR{$T / T_p$ passes}
    \STATE Select subset, $X_p = s(f_\theta, X, k)$
    \FOR{$T_p$ epochs}
        \STATE Train $f_\theta$ on $X_p$
    \ENDFOR
\ENDFOR
\end{algorithmic}
\end{algorithm}

\subsection{Pruning selection criterion}

In the next subsections, we describe the different selection criteria, $s$, that are proposed in our work.

\subsubsection{Uncertainty sampling with EMA}

Uncertainty sampling preferentially selects samples for which the model's current prediction is least confident. In general, it is quick to compute and identifies points close to the estimated decision boundary which is a measure of the informativeness of a sample. Specifically, we use the per-sample cross entropy loss for assigning a score to individual sample $x$ with label $y$ over $C$ classes:

\begin{equation}
    s_{\text{uncertainty}}(x) = \sum_{i = 1}^C -y_i \log \left(\frac{\exp(f_\theta(x_i))}{\sum_{j = 1}^C \exp(f_\theta(x_j))}\right)
    \label{eq:uncertainty}
\end{equation}

This scoring criterion is similar to the EL2N score from Paul \etal \cite{paul_deep_2021}. In that work, the final EL2N score is obtained by averaging the uncertainty over several trials to reduce the noise given in a single trial. We observe the same noisy behavior across pruning checkpoints in our work. A network's decision boundary can vary dramatically, especially with respect to points which are included in one checkpoint and excluded in another. Since we are pruning dynamically, instead of a static average, we use an exponential moving average filter:

\begin{equation}
    s_{\text{ema}}(x) = \alpha s_{\text{uncertainty}}(x) + (1 - \alpha) s_{\text{previous ema}}(x)
    \label{eq:uncertainty-ema}
\end{equation}
where $s_{\text{previous ema}}(x)$ is the value of $s_{\text{ema}}(x)$ from the previous checkpoint.

\subsubsection{$\epsilon$-greedy approach} 

So far, our decision making criterion has been fairly simple. The points with the highest average loss are selected at each checkpoint. This approach can be viewed as a greedy solution to a multi-armed bandit problem from reinforcement learning \cite{sutton_reinforcment_2018}. Each sample is an arm associated with some unknown value (i.e. score). By selecting points, we make $k$ pulls of various arms, and when we measure the uncertainty on the next checkpoint, we make a noisy observation of the value of every sample in the dataset.

Selecting the samples with the highest uncertainty at every checkpoint corresponds to a greedy decision. Since we maximally exploit our current information, we might under-select some points and obtain a suboptimal solution. A simple fix is to encourage more exploration by randomly selecting points which do not currently have the highest score. This is referred to as an $\epsilon$-greedy approach in the bandit setting \cite{sutton_reinforcment_2018}.

Concretely, at every frequency checkpoint, $(1 - \epsilon) k$ points are selected using \cref{eq:uncertainty-ema}, and $\epsilon k$ points are selected from the remaining data uniformly at random. Our results show that this additional exploration can help boost performance even further.

\subsubsection{Upper-confidence bound approach} 

While the $\epsilon$-greedy approach does introduce more exploration, it does so in a random manner. A more directed approach would be to prioritize samples for which the running variance of \cref{eq:uncertainty} is high even when the mean is low. In this way, when we do explore, we choose to focus on the samples who's scores are poorly estimated.

The upper-confidence bound (UCB) \cite{sutton_reinforcment_2018} algorithm is another RL method that selects arms based on the mean value and the variance. Assuming the initial mean scores are aggregated from the random model and the starting variance is zero, we compute the variance based on Welford's algorithm \cite{welford_note_1962}:
\begin{align}
    \mathrm{var}(x) &= (1 - \alpha) \mathrm{var}_{\text{previous}}(x) \\ \nonumber
    &\quad + \alpha (s_\text{uncertainty}(x) - s_\text{previous ema}(x))^{2}
    \label{eq:ucb-var}
\end{align}
With the mean and variance computed, the final score of the sample is computed by adding the two quantities together:
\begin{equation}
    s_{\text{ucb}}(x) = s_{\text{ema}}(x) + c \mathrm{var}(x)
    \label{eq:ucb}
\end{equation}
where $c$ is a hyper-parameter the variance's influence on the total score.
\section{Results and discussion}

We experimentally evaluate our methods on the CIFAR-10 and CIFAR-100 datasets \cite{krizhevsky_learning_2009} (following the setup in Paul \etal \cite{paul_deep_2021}). Each image is augmented with standard transformations such as normalizing the images by per channel mean and standard deviation, padding by 4 pixels, random cropping to 32 by 32 pixels, and horizontal flipping with probability 0.5. We use ResNet-18 \cite{he_deep_2016} with the initial 7x7 convolutional layer and 2x2 pooling layer swapped out for a 3x3 convolution layer and a 1x1 pooling layer to account for change in image resolution. We train our model for 200 epochs with the stochastic gradient descent (SGD) optimizer with an initial learning rate = 0.1 with Nesterov momentum = 0.9, weight decay = 0.0005, and batch size = 128. The learning rate was decayed by a factor of 5 at 60, 120, and 160 epochs. Our implementation is based on PyTorch \cite{paszke_automatic_2017} and all experiments were conducted on a cluster of 11GB NVIDIA GeForce RTX 2080 Ti GPUs. For CIFAR100, we use the same experimental setup but swap out ResNet-18 for ResNet-34. All our code is available in the supplementary material (and will be public once published).

For our baselines, we compare against conventional training with the full dataset, forget scores, and EL2N scores. For the forget scores, we train the model for 200 epochs and compute the forget scores as per the prior work. For the EL2N scores, we also follow the scoring algorithm laid out in the prior work and average over 10 independent models trained for 20 epochs each. After obtaining the forget and EL2N scores, we statically prune the data set by selecting the top scoring samples based on the designated pruning rate. For all our experiments, we run each method over 4 independent random seeds.

\subsection{CIFAR-10 results}

\begin{figure}[!h]
    \centering
    \begin{subfigure}{0.45\textwidth}
        \centering
        \includegraphics[width=\textwidth]{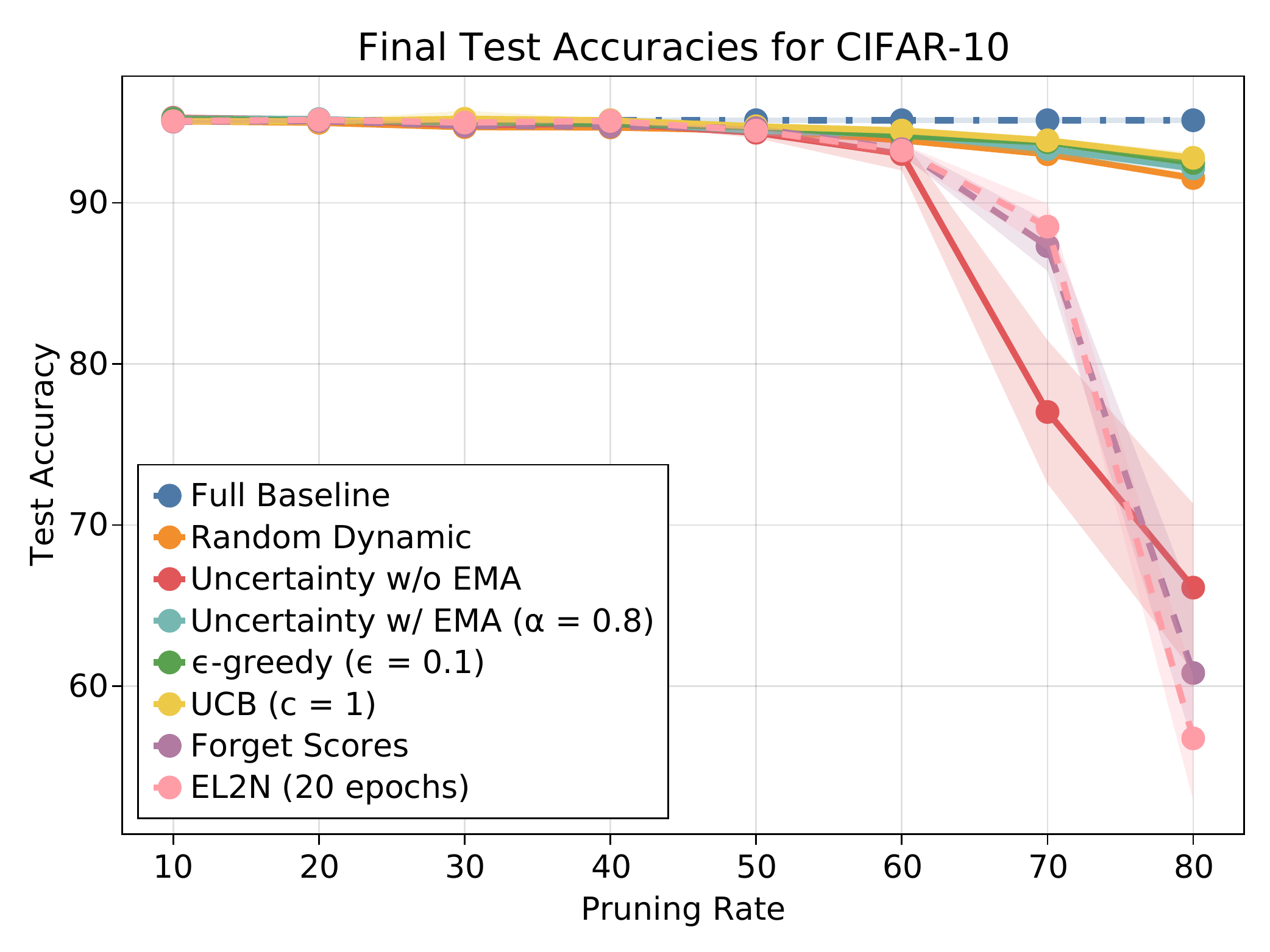}
        \caption{CIFAR10 accuracy}
        \label{fig:cifar10_acc}
    \end{subfigure}
    \begin{subfigure}{0.45\textwidth}
        \centering
        \includegraphics[width=\textwidth]{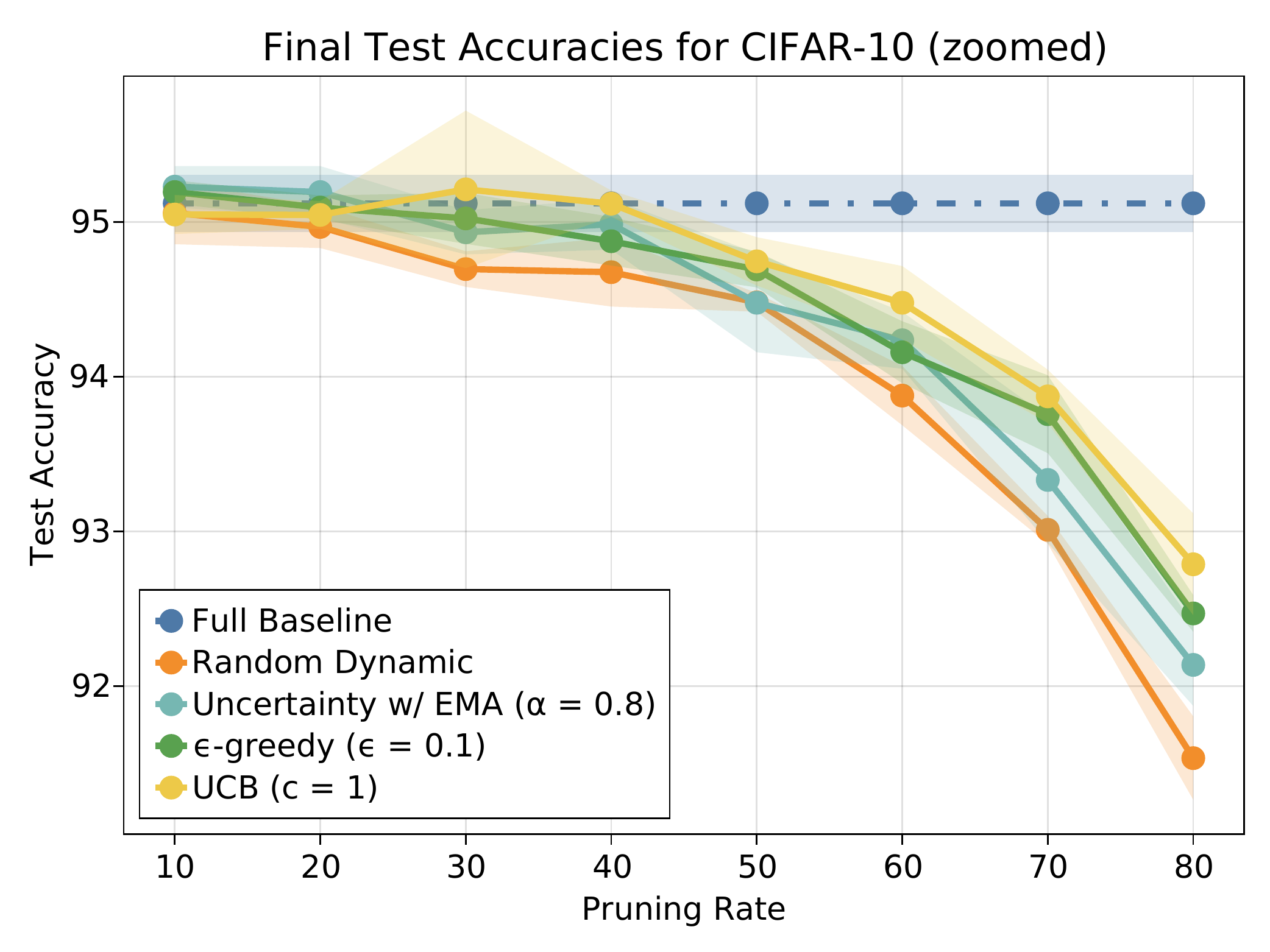}
        \caption{CIFAR10 accuracy (zoomed)}
        \label{fig:cifar10_acc_zoom}
    \end{subfigure}
    \begin{subfigure}{0.45\textwidth}
        \centering
        \includegraphics[width=\textwidth]{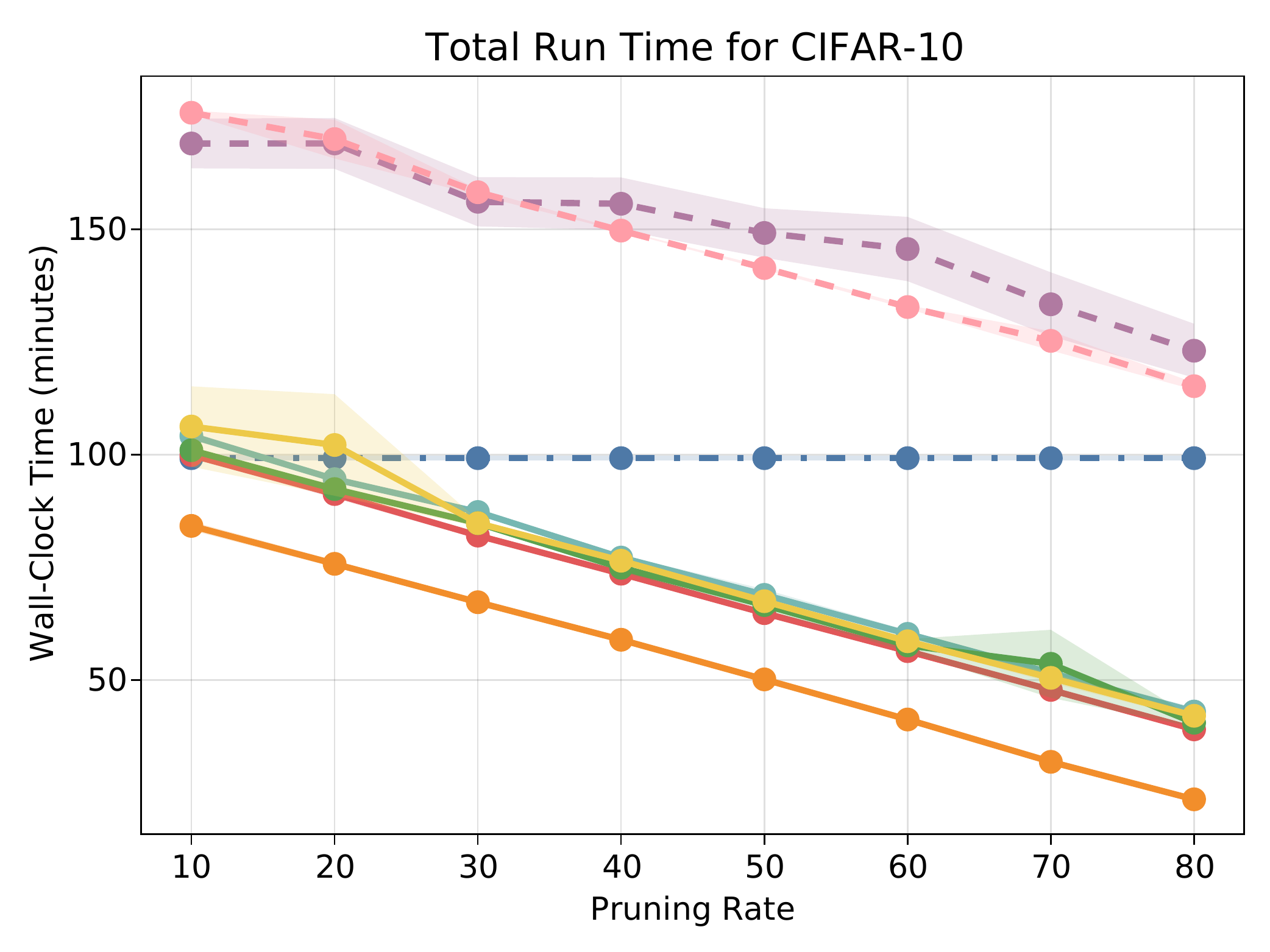}
        \caption{CIFAR10 run time}
        \label{fig:cifar10_wc}
    \end{subfigure}
    \caption{Dynamic data pruning applied to CIFAR-10 with ResNet-18. \cref{fig:cifar10_acc} shows the final test accuracies for each method, \cref{fig:cifar10_acc_zoom} shows the same data without the under-performing methods, and \cref{fig:cifar10_wc} shows the run time to execute the approach (including scoring cost).}
    \label{fig:cifar10_results}
\end{figure}

\cref{fig:cifar10_results} shows the result of dynamic data pruning with the various selection methods introduced in the prior section. We sweep the pruning rate with the pruning period ($T_p$) set to every 10 epochs. For all methods that use an EMA, we set $\alpha = 0.8$. 

Below 50\% of the data pruned, all methods appear to perform equally well. At more aggressive rates, the static baselines and the uncertainty without the EMA drop dramatically in performance. Our proposed methods continue to maintain reasonable performance even when 80\% of the dataset is pruned. As expected, the $\epsilon$-greedy and UCB improve the performance slightly over the uncertainty with the EMA.

In \cref{fig:cifar10_wc}, we plot the measured wall-clock time for each method. Since the static methods perform pruning prior to training with an overhead equivalent to training on the full data, their wall-clock time is nearly double the full baseline (when omitting the static overhead, the wall-clock time is on par with random pruning). In contrast, all the dynamic methods reduce the wall-clock time even at modest pruning rates.

\subsection{Unreasonable effectiveness of random pruning}

The most surprising result in \cref{fig:cifar10_acc} is the performance of the random dynamic pruning algorithm. This method randomly samples the training data at each checkpoint without any consideration to the uncertainty or previous checkpoints. Still, the approach is able to outperform the static baselines, and it suffers only a 3.5\% drop in accuracy at a pruning rate of 80\%. While this is unexpected in the context of this paper, it does match intuition from first principles. The very basis of SGD training relies on the ability to train on randomly sampled batches. If there was a strong bias in the relative importance of samples, then most DL training would be quite inefficient.

\begin{figure}
    \centering
    \includegraphics[width=0.45\textwidth]{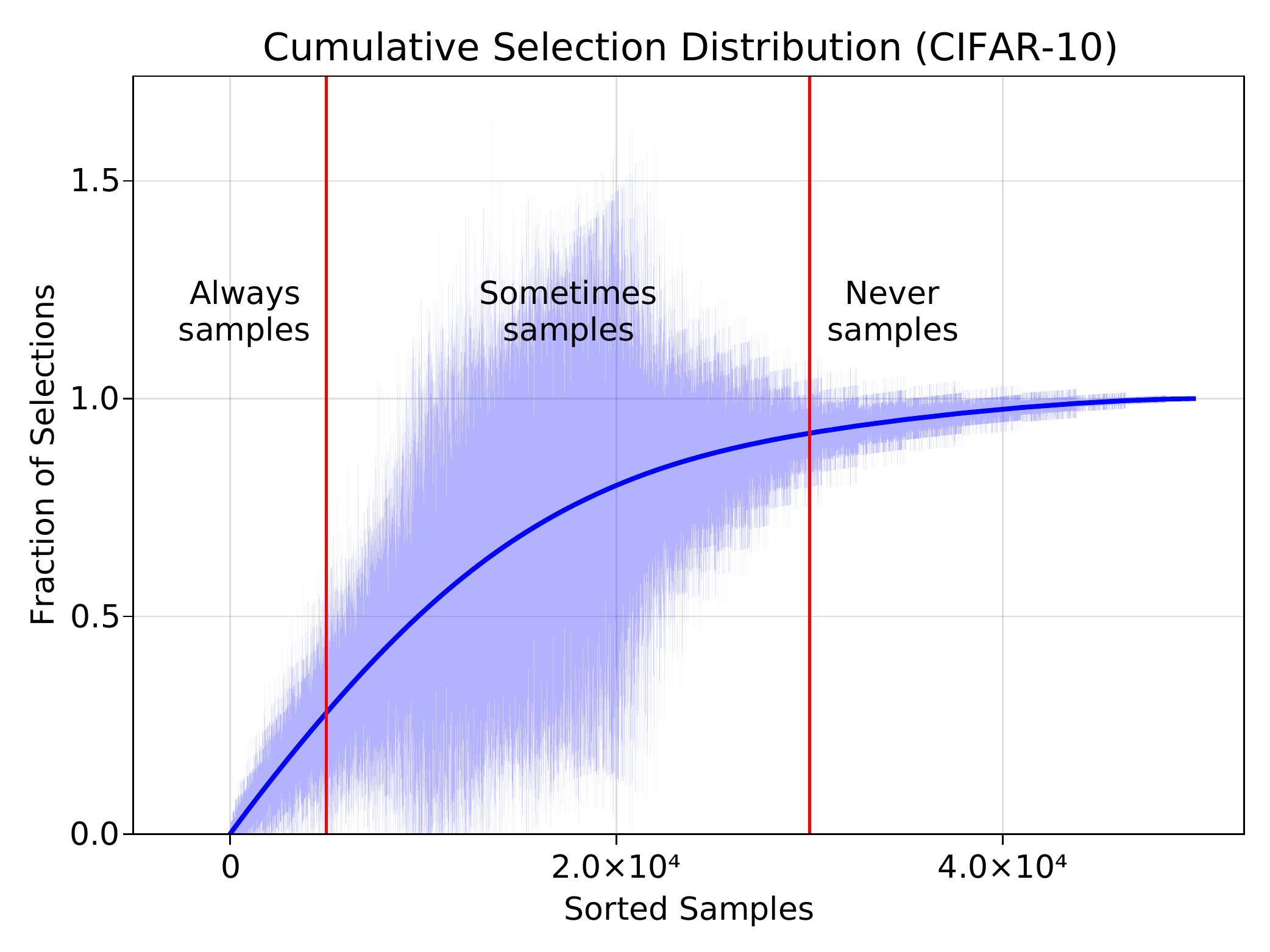}
    \caption{The cumulative fraction of times each sample from CIFAR-10 is selected during our uncertainty with EMA method at 70\% pruning. The horizontal axis is sorted from most frequently selected to least frequently selected. The data is separated into three groups --- samples that are always selected, samples that are selected only some of the time, and samples that are never selected. Static methods fail to effectively target the middle group.}
    \label{fig:cifar10-score-cdf}
\end{figure}

This is corroborated by the our findings on the existence of ``sometimes'' points. When performing each of the trials to obtain \cref{fig:cifar10_results}, we count the number of times each sample is selected across all checkpoints. This count is normalized by the total number of ``slots'' available across all checkpoints. Thus, we obtain the fraction of times a given sample was selected out of all possible opportunities for selection in a single trial. We plot the mean and standard deviation across trials of this distribution in \cref{fig:cifar10-score-cdf}. The horizontal axis is sorted so that the most frequently selected samples are on the left, and the least frequently selected are on the right.

We observe that the curve can be qualitatively separated into three distinct regions --- always, sometimes, and never samples. Always samples are selected at nearly every checkpoint indicated by how steeply the cumulative distribution increases initially. Moreover, the standard deviation in this region is low, meaning that the always samples are consistent across trials. Similarly, the never samples are rarely selected, resulting in a slow rate of increase at the end of the curve. Like always samples, these points are consistently never selected.

Contrary to always and never samples, sometimes samples are not consistently selected or not. The rate of increase in the curve slows down as each sometimes sample contributes a smaller fraction of the total selections. Moreover, the high variance across trials in this region shows that the ranking of sometimes samples is not consistent. This suggests that a large portion of the dataset is composed of samples that are difficult to rank.

Static pruning methods largely target the elimination of never samples. By selecting the top $k$ points in a single pass, static methods suffer from the variability of sometimes samples. In contrast, dynamic methods can rotate their selection of sometimes samples based on which ones are most beneficial to a specific model's training trajectory. Most importantly, since such a large fraction of the dataset does not have a consistent ranking, even randomly sampling these points (as the random dynamic method does) is better than statically committing to a fixed subset of them.

\begin{table}[!hb]
  \begin{center}
    \caption{The effect of randomly selecting sometimes samples when re-training a model using previous scores.}
    \label{tab:retrain}
    \begin{tabular}{lr}
        \toprule
        \textbf{Method} & \textbf{Accuracy} \\ \midrule
        Original dynamic training & 93.33\% \\
        Always set + static sometimes set & 89.52\% \\
        Always set + random sometimes set & 92.50\% \\ \bottomrule
    \end{tabular}
  \end{center}
\end{table}

We test this hypothesis by training a model on CIFAR-10 using the uncertainty with EMA method at a pruning rate of 70\%, and we save the scores across checkpoints. Then, we train a new initialization with dynamic pruning using two different subsets of the original dataset. First, we choose the top 30\% of samples according to the scores from the final checkpoint. This creates a static policy that uses the same subset at every checkpoint. Second, we choose only the always samples, which make up about 15\% of the data, and we choose the remaining samples in the pruning budget uniformly at random for each checkpoint. The difference is shown in \cref{tab:retrain}. Statically choosing the sometimes samples results nearly a 4\% performance drop. Dynamically sampling from the remaining samples only results in a 1\% performance degradation.

\subsection{CIFAR-100 results}

\begin{figure}[!ht]
    \centering
    \begin{subfigure}{0.45\textwidth}
        \centering
        \includegraphics[width=\textwidth]{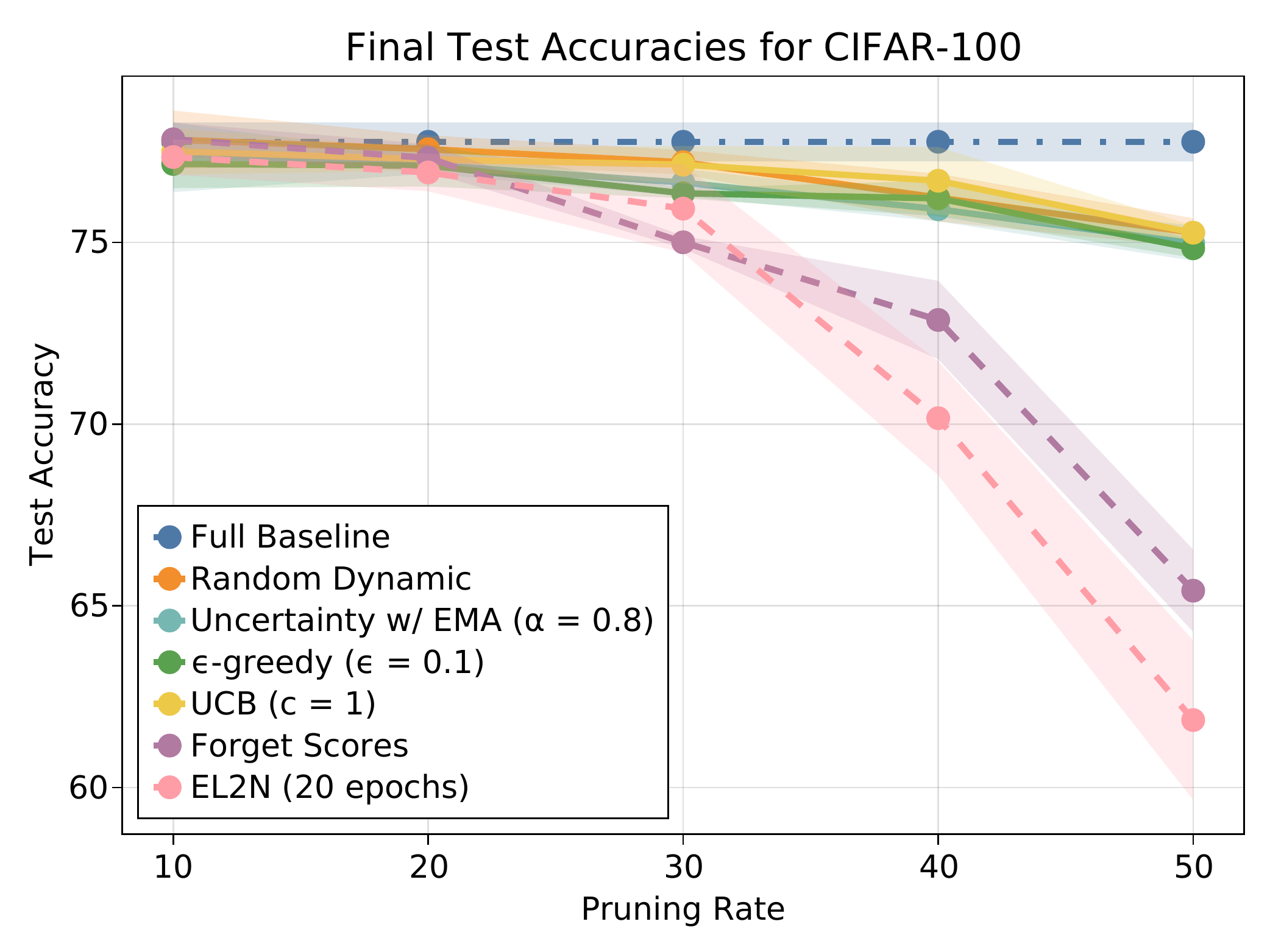}
        \caption{CIFAR100 accuracy}
        \label{fig:cifar100_acc}
    \end{subfigure}
    \begin{subfigure}{0.45\textwidth}
        \centering
        \includegraphics[width=\textwidth]{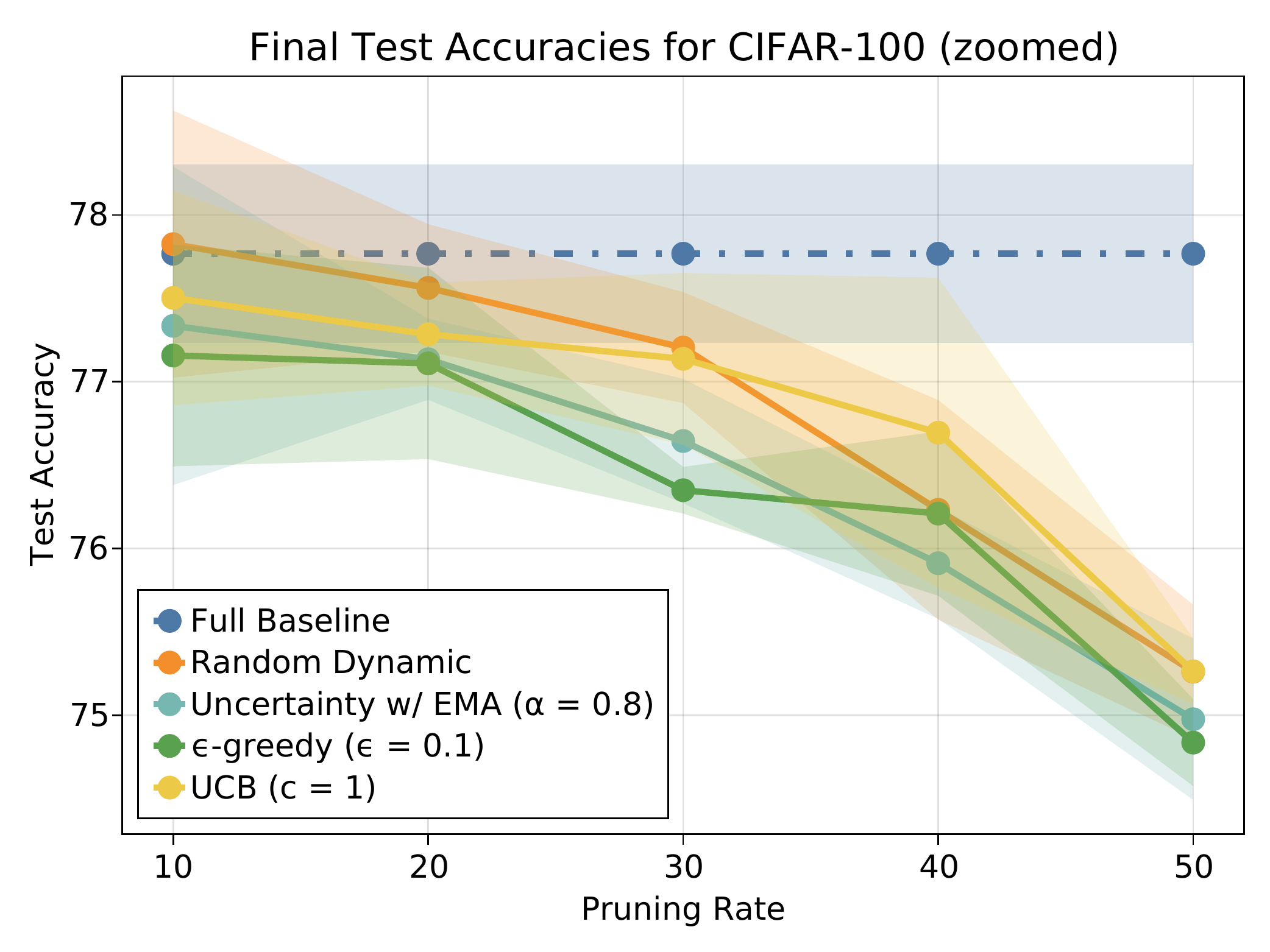}
        \caption{CIFAR100 accuracy (zoomed)}
        \label{fig:cifar100_acc_zoom}
    \end{subfigure}
    \begin{subfigure}{0.45\textwidth}
        \centering
        \includegraphics[width=\textwidth]{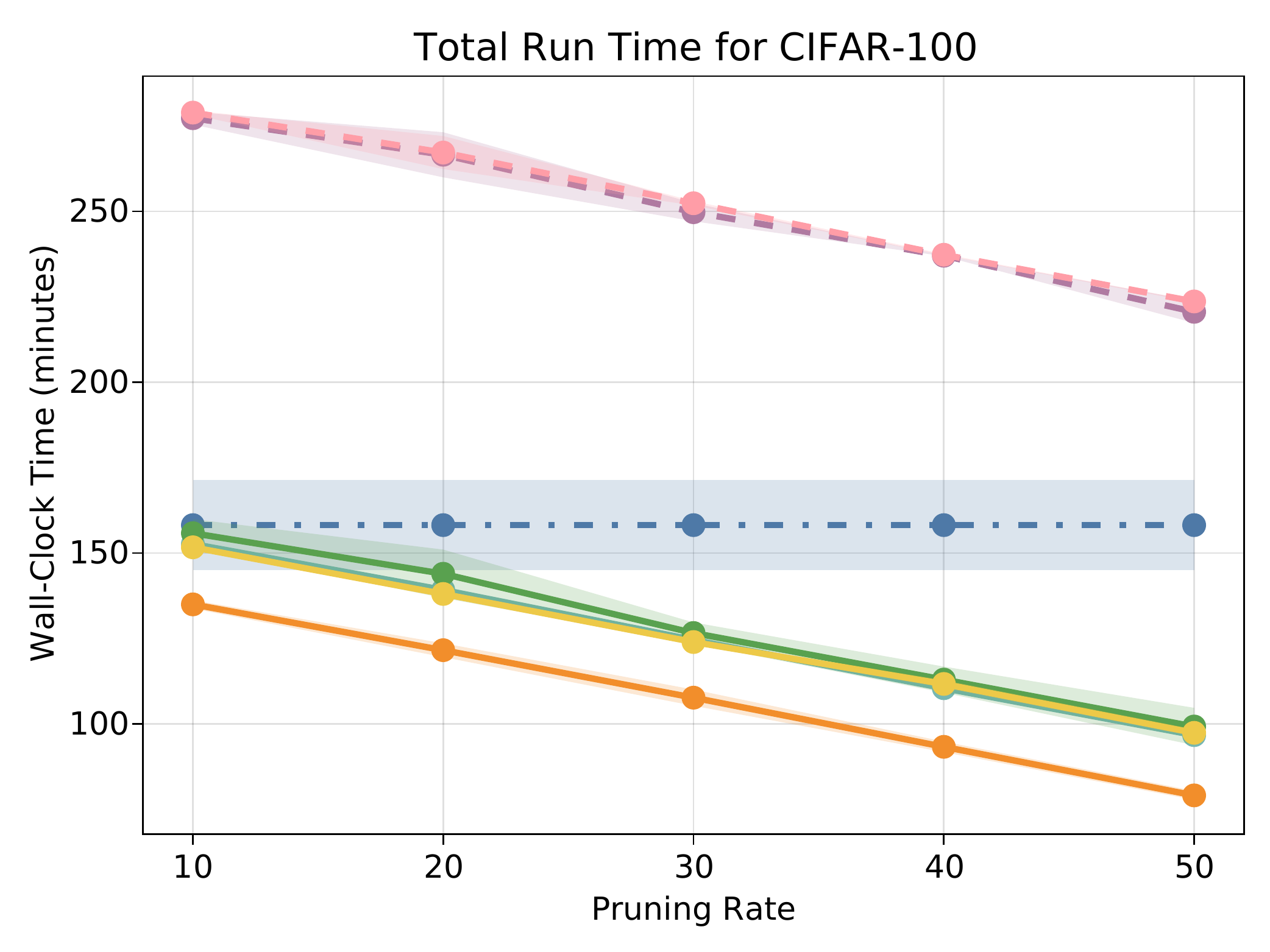}
        \caption{CIFAR100 run time}
        \label{fig:cifar100_wc}
    \end{subfigure}
    \caption{Dynamic data pruning CIFAR-100 with ResNet-34. \cref{fig:cifar100_acc} shows the final test accuracies for each method, \cref{fig:cifar100_acc_zoom} shows the same data without the static methods, and \cref{fig:cifar100_wc} shows the run time to execute the approach (including scoring cost).}
    \label{fig:cifar100_results}
\end{figure}

\cref{fig:cifar100_results} shows the result of dynamic data pruning on CIFAR-100. We retained the same hyper-parameters as in the CIFAR-10 result but omit the uncertainty method without the EMA as its performance was unsatisfactory. We can see that EL2N and forget scores are able to nearly retain the performance up to a pruning rate of 20\% but deteriorate rapidly beyond that point.

\begin{figure}[!hb]
    \centering
    \includegraphics[width=0.45\textwidth]{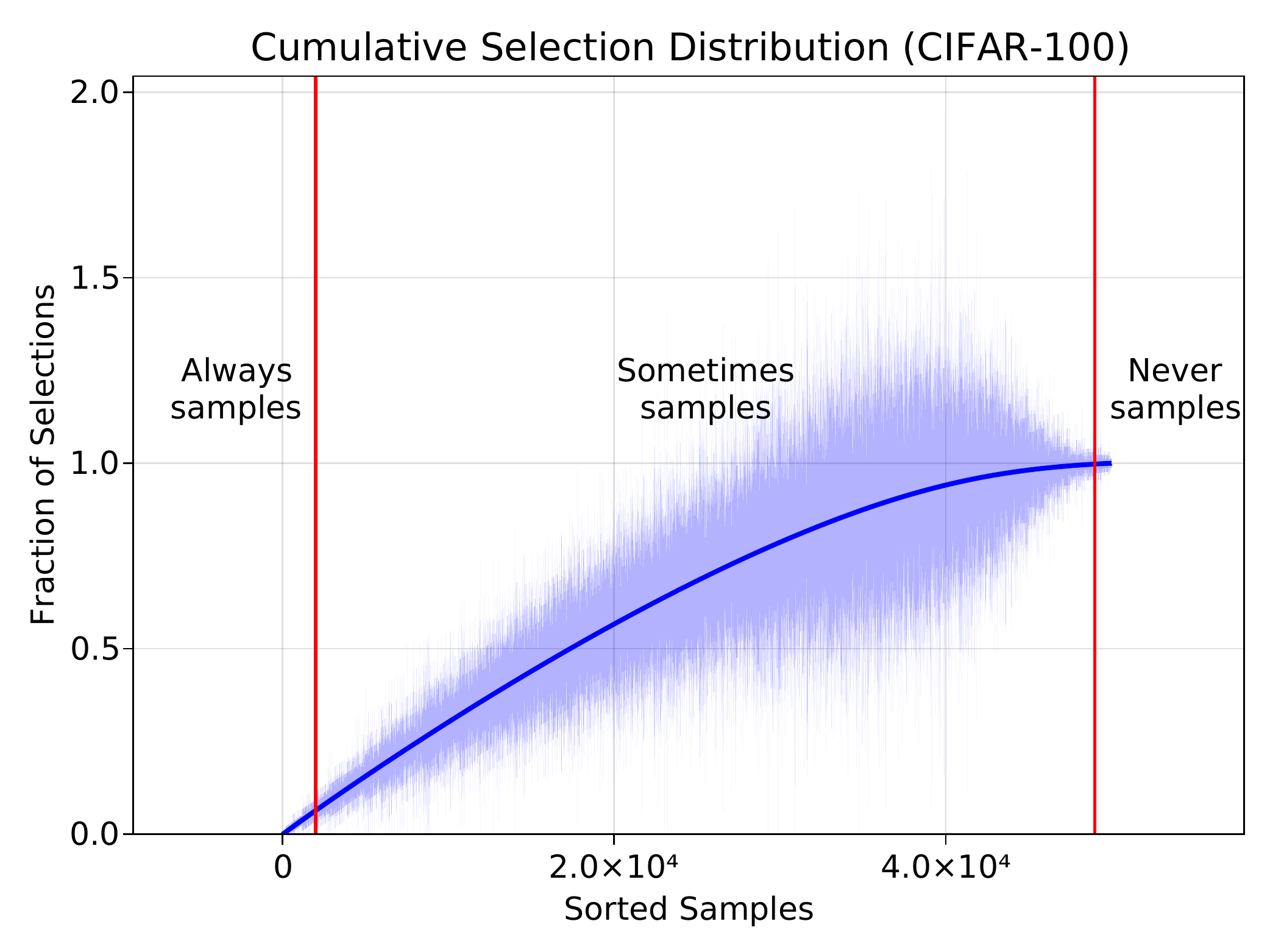}
    \caption{Cumulative selection distribution of samples on CIFAR-100 at 40\% pruning with UCB. The horizontal axis is sorted with the most frequently selected samples on the left. Unlike CIFAR-10, CIFAR-100 has very few always and never samples. Thus, a random dynamic method can prune close to optimally on this dataset.}
    \label{fig:cifar100-score-cdf}
\end{figure}

All the dynamic methods outperform the static baselines. While the UCB does appear to have slightly higher accuracy than the random dynamic method at more aggressive pruning rates, most of our methods underperform compared to the random dynamic method. This is not surprising when we consider the cumulative selection distribution in \cref{fig:cifar100-score-cdf}. Unlike CIFAR-10, CIFAR-100 has almost no always/never points. This indicates that there is very little opportunity to select samples more intelligently than a random dynamic method. One plausible explanation for the lack of always/never samples is the low number of samples per class in CIFAR-100 (500 vs. 5000 for CIFAR-10). Thus, each sample has more influence on the decision boundary, and the intrinsic redundancy in the dataset is much lower.

\subsection{Influence of hyperparameters}
Our methods have several hyper-parameters that can be adjusted --- the pruning rate ($k / N$), the pruning period ($T_p$), the exploration rate ($\epsilon$; $\epsilon$-greedy only), and the confidence ($c$; UCB only). Our previous results fix most of these hyper-parameters and vary only the pruning rate. In this section, we explore the effect of the remaining hyper-parameters. The overall conclusion is that the remaining hyper-parameters have minimal effect on the end result. The hyper-parameter sweeps are reported in the supplementary material.

\subsection{Imbalanced CIFAR-10 results}

So far, we have studied standard datasets which are well balanced. Our CIFAR-100 results suggest that the number of samples per class can have a large influence on the pruning opportunity. To further explore this phenomenon, we generate a synthetically imbalanced variant of CIFAR-10 by sub-sampling each class according to \cref{tab:imbalance}.

\begin{table}[t]
  \begin{center}
    \caption{The subsampling rates by class for the imbalanced CIFAR-10 dataset.}
    \label{tab:imbalance}
    \begin{tabular}{lr}
        \toprule
        \textbf{Class Indices} & \textbf{Subsample Rate} \\ \midrule
        0, 1 & 25\% \\
        2, 3, 4 & 50\% \\
        5, 6 & 75\% \\
        7, 8, 9 & 100\% \\ \bottomrule
    \end{tabular}
  \end{center}
\end{table}

\begin{figure}[!ht]
    \centering
    \includegraphics[width=0.45\textwidth]{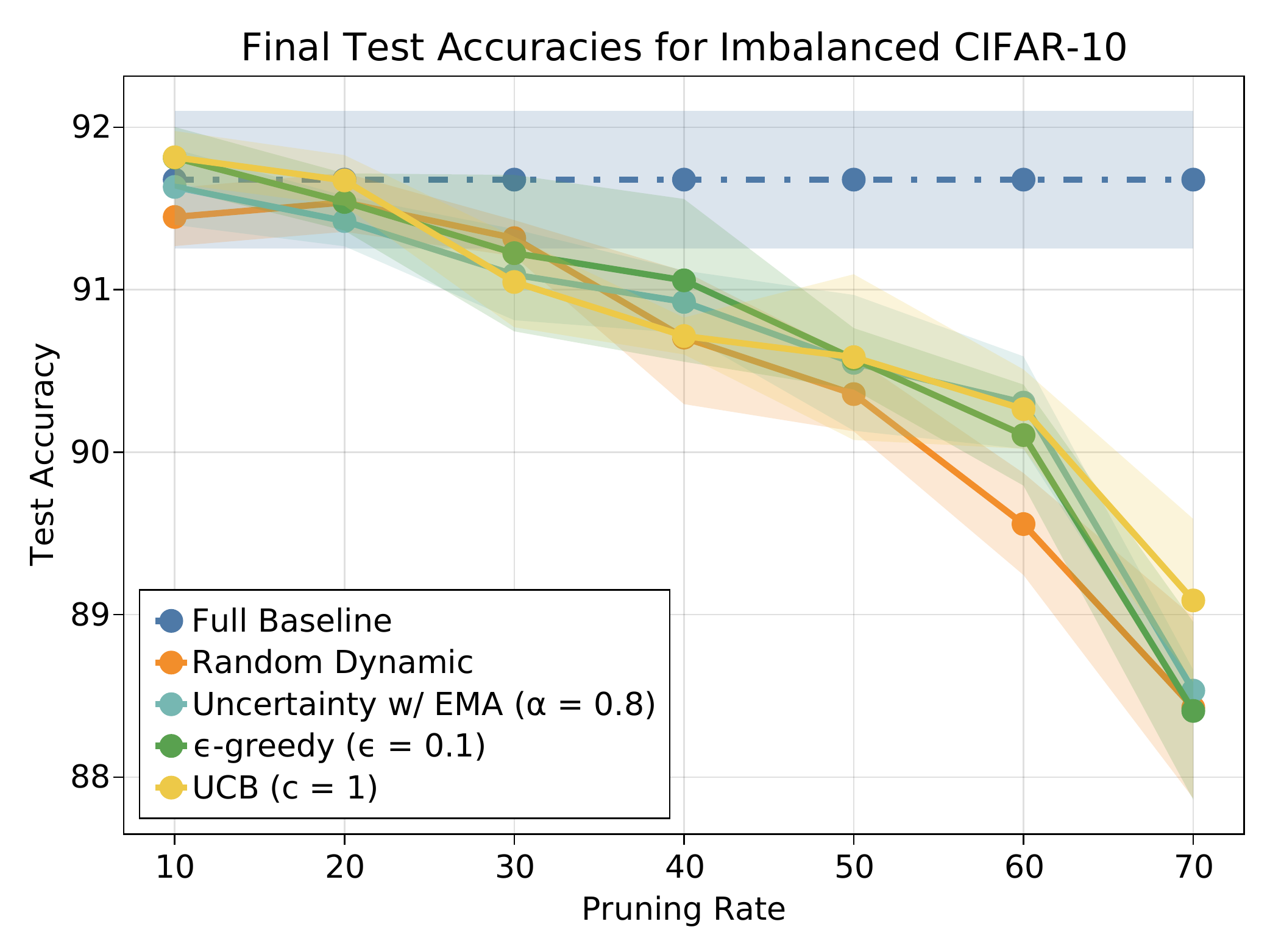}
    \caption{Dynamic data pruning a synthetically imbalanced CIFAR-10 with ResNet-18. The horizontal axis refers to the amount of data removed from the training set, and the vertical axis shows the final test accuracies for each method.}
    \label{fig:imbalance-cifar-10-accuracy}
\end{figure}

\begin{figure}
    \centering
    \includegraphics[width=0.45\textwidth]{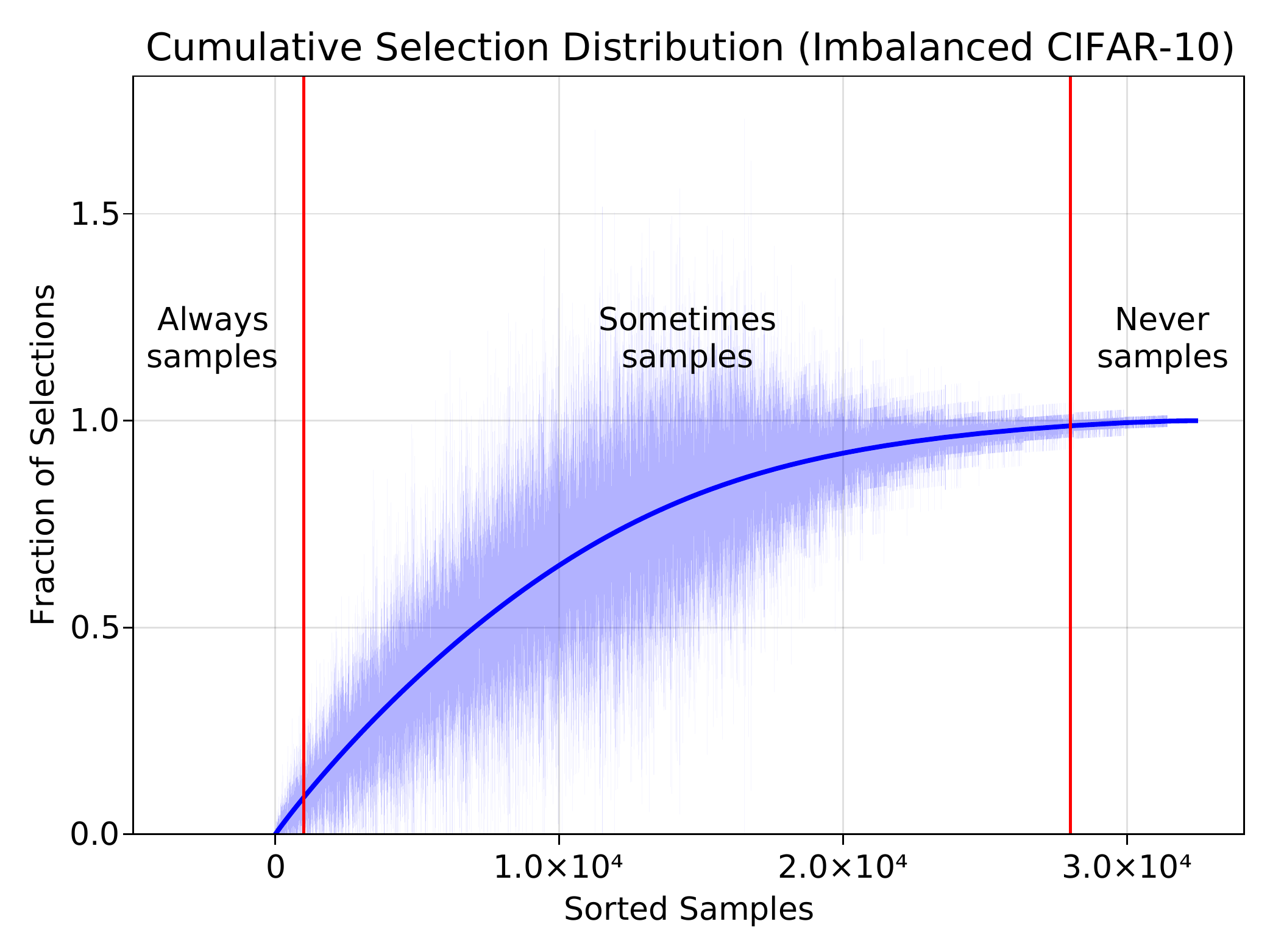}
    \caption{Cumulative selection distribution of samples on the imbalanced CIFAR-10 dataset at 40\% pruning with UCB. The horizontal axis is sorted with the most frequently selected samples on the left.}
    \label{fig:imbalance-cifar-10-score-breakdown}
\end{figure}

The results are shown in \cref{fig:imbalance-cifar-10-accuracy}. Our methods outperform the random dynamic method at aggressive pruning rates, but all methods lose performance at modest pruning rates. This is expected based on \cref{fig:imbalance-cifar-10-score-breakdown}. The lack of always/never samples makes this dataset difficult to prune. Unlike CIFAR-100, there does appear to be some slowdown in the rate of increase along the cumulative curve. This indicates that there is some coarse ordering to the sometimes samples that can be exploited. The UCB method is able to target this opportunity once the pruning rate is $> 50\%$.

\subsection{Downsampled CIFAR-10 results}

Similar to the imbalanced CIFAR-10 dataset, we also study a downsampled variant of CIFAR-10. We decrease the number of samples per class from 5000 to 1000 for every class. This mimics CIFAR-100 where the number of samples per class is 500.

\begin{figure}[!ht]
    \centering
    \includegraphics[width=0.45\textwidth]{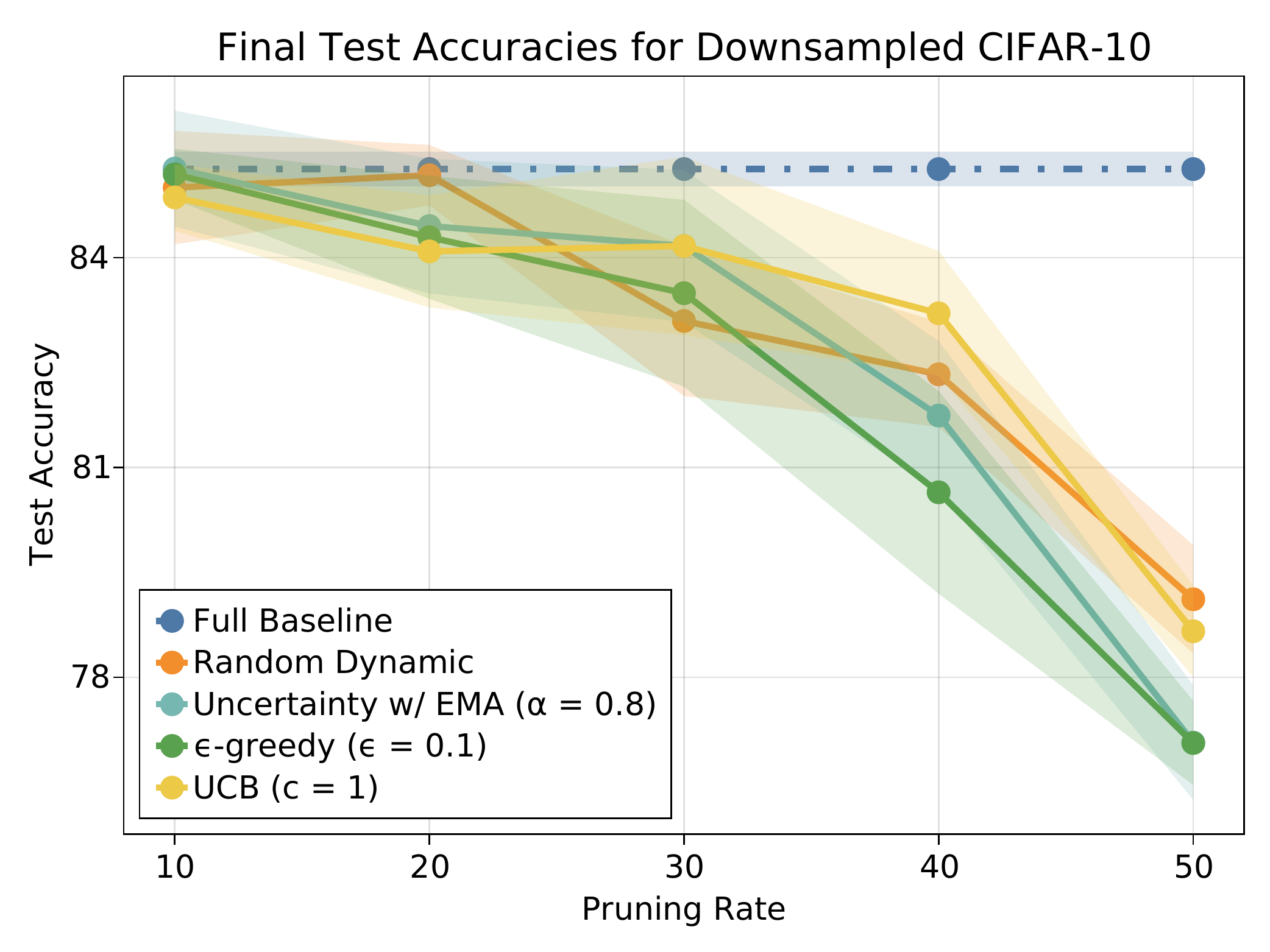}
    \caption{Dynamic data pruning a synthetically downsampled CIFAR-10 with ResNet-18. The horizontal axis refers to the amount of data removed from the training set, and the vertical axis shows the final test accuracies for each method.}
    \label{fig:downsample-cifar-10-accuracy}
\end{figure}

\begin{figure}[!ht]
    \centering
    \includegraphics[width=0.45\textwidth]{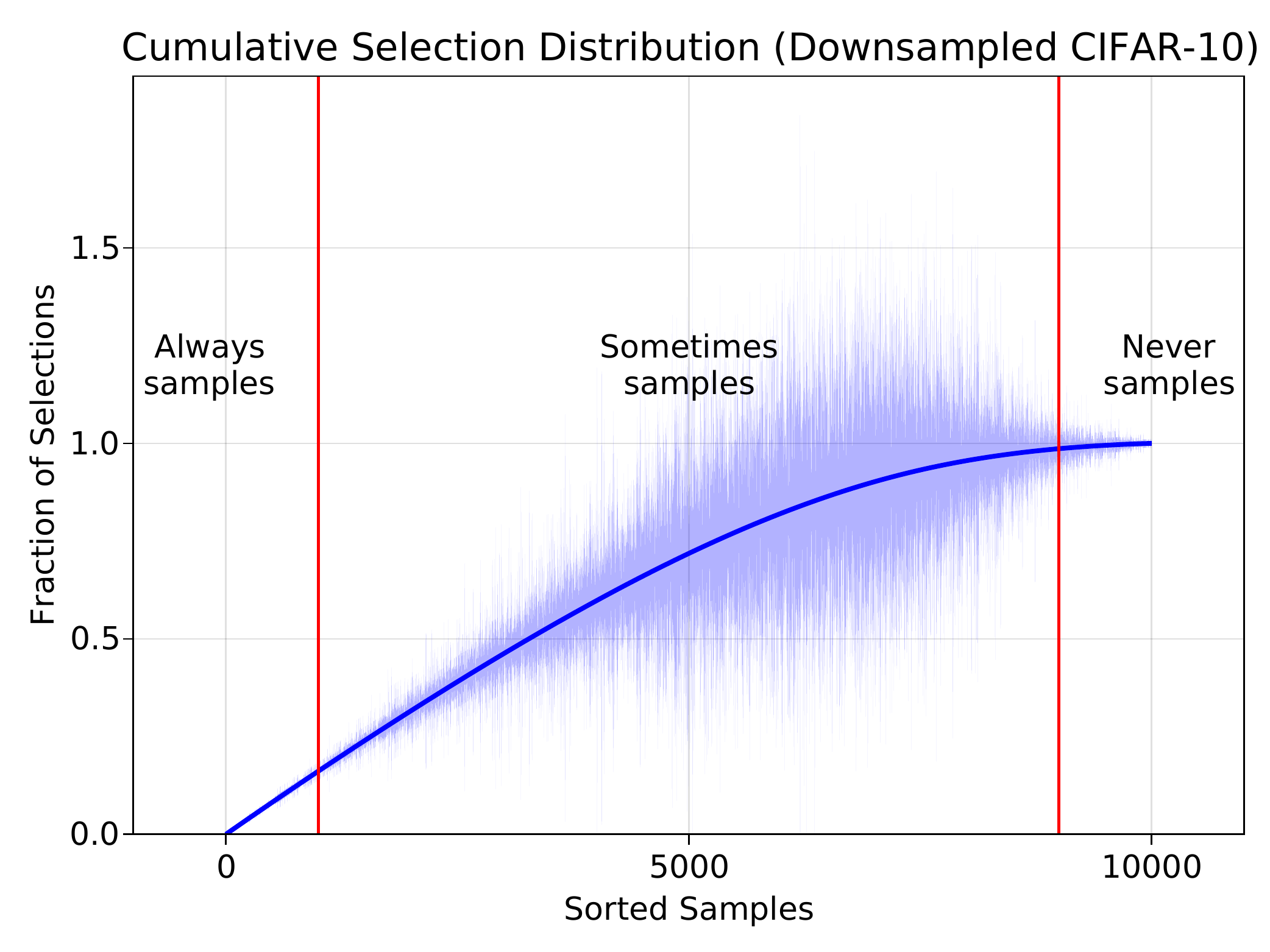}
    \caption{Cumulative selection distribution of samples on the downsampled CIFAR-10 dataset at 40\% pruning with UCB. The horizontal axis is sorted with the most frequently selected samples on the left.}
    \label{fig:downsample-cifar-10-score-breakdown}
\end{figure}

The results are shown in \cref{fig:downsample-cifar-10-accuracy}. Similar to the CIFAR-100 results, the uncertainty with EMA and $\epsilon$-greedy methods are worse than the random dynamic method. Only the UCB method matches the random dynamic method in performance. These results are corroborated by \cref{fig:downsample-cifar-10-score-breakdown} which shows that the sample selection distribution follows the same trend as CIFAR-100.

\section{Conclusion}

This work presents a dynamic approach to data pruning. We reframe the problem as a decision making process, and we present three methods (uncertainty with EMA, $\epsilon$-greedy, UCB) for pruning datasets online while training. This allows our methods to be applied to novel datasets that are being trained on for the first time. Unlike the prior work, our approaches do not incur significant overhead, and we are able to reduce the total run time.

Moreover, we introduce the notion of a cumulative selection distribution which separates a dataset into three groups --- always samples, sometimes samples, and never samples. We show that sometimes samples are difficult to rank, and as a result, static pruning methods cannot effectively select a subset of them. Surprisingly, we find that some datasets, like CIFAR-100, have little opportunity for sophisticated pruning. This is because the dataset has very few always/never samples. In these cases, the best option is to prune the dataset using a random dynamic method, since all samples have equal importance. We corroborate these conclusions by testing these hypotheses on synthetically modified CIFAR-10 datasets with severe class imbalance or low samples-per-class.

We hope our work emphasizes the need to understand data pruning as a function of the dataset and the model. By viewing the pruning problem as an online decision-making process, we expect future work to borrow from active learning and reinforcement learning to more effectively target sometimes samples. In lieu of these improvements, our methods bring practical, efficient data pruning to DL researchers.

\subsection{Societal impacts}

This work has financial and environmental societal implications. Current trends in deep learning have made training state-of-the-art models prohibitively expensive. As a result, only large research organizations have access to large, overparameterized models. By reducing the total training time, our methods allow independent researchers to train more sophisticated models. Moreover, unlike the prior work, our methods can be applied directly to novel datasets.

More importantly, large-scale deep learning requires weeks of compute time on an energy-hungry GPU cluster. Many of these GPUs are used to parallelize iterating over the dataset. By pruning the data, we allow models to be trained with fewer GPUs or in less time. Both outcomes translate to lower energy consumption and carbon footprints for DL training.

On the other hand, by helping to democratize complex DL models to new domains, our work may negatively impact areas of society affected by those domains. Namely, since bias and robustness of DL is poorly understood, an increase in its applicability may cause unexpected harm.

\subsection{Limitations and future work}

Our extensive experiments show that dynamic data pruning is an effective methodology to accelerate deep learning training; however, there are limitations to our techniques. As the proposed methods rely upon the per-sample loss, label noise would hurt the accuracy of the classifier significantly more than the full baseline as each individual sample has more influence on the decision boundary.

Additionally, our RL-based methods fail to outperform a random dynamic baseline for datasets with many sometimes samples, like CIFAR-100. In these contexts, a more thorough understanding of sometimes samples could lead to a more sophisticated approach that improves over random. In particular, we separate the dataset into the three regions qualitatively. A more principled, formal approach to labeling each samples as ``always,'' ``sometimes,'' or ``never'' could help tailor pruning methods to the specific dataset.

Moreover, our methods only make a decision about which samples to keep at each checkpoint, but future methods could include various hyperparameters as part of the decision making process. For example, an algorithm could decide the samples to keep at the current checkpoint, as well as the period until the next checkpoint or the pruning rate at the current checkpoint. Such methods could iteratively prune the dataset leading to a better final test accuracy.
{\small
\bibliographystyle{ieee_fullname}
\bibliography{ref}
}
\newpage
\appendix
In the supplementary material below, we present additional results showing a more refined run time breakdown for the static prior work, an application of our RL techniques to the static methods, and a complete hyperparameter sweep for our methods. We also provide details on how to reproduce our results using the packaged source code.

\section{Run time breakdown for static methods}
\begin{figure}
    \centering
    \includegraphics[width=0.45\textwidth]{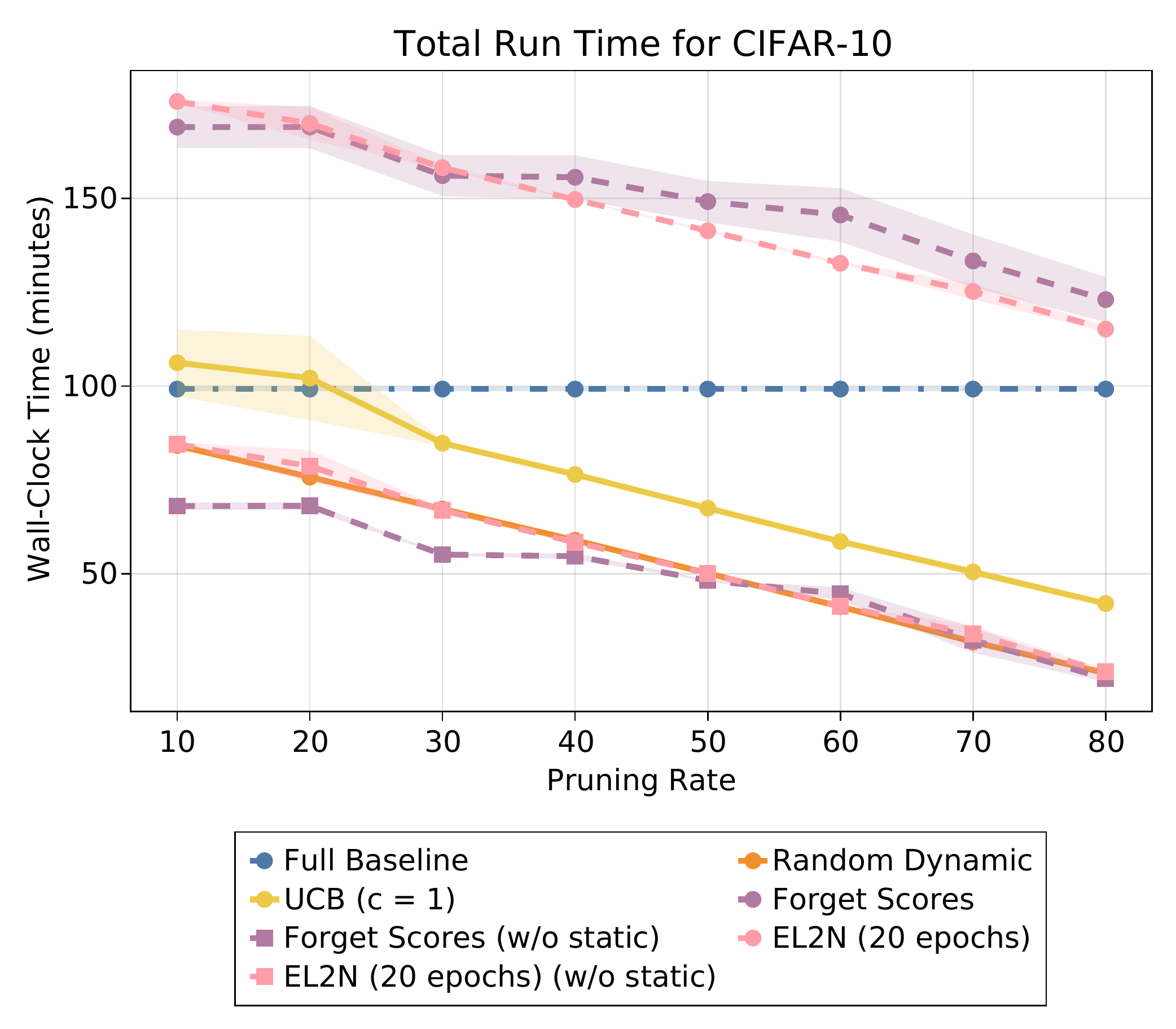}
    \caption{Dynamic data pruning on CIFAR-10 with pre-training scoring cost included and excluded for static methods.}
    \label{fig:static_cifar10_wc}
\end{figure}

\begin{figure}
    \centering
    \includegraphics[width=0.45\textwidth]{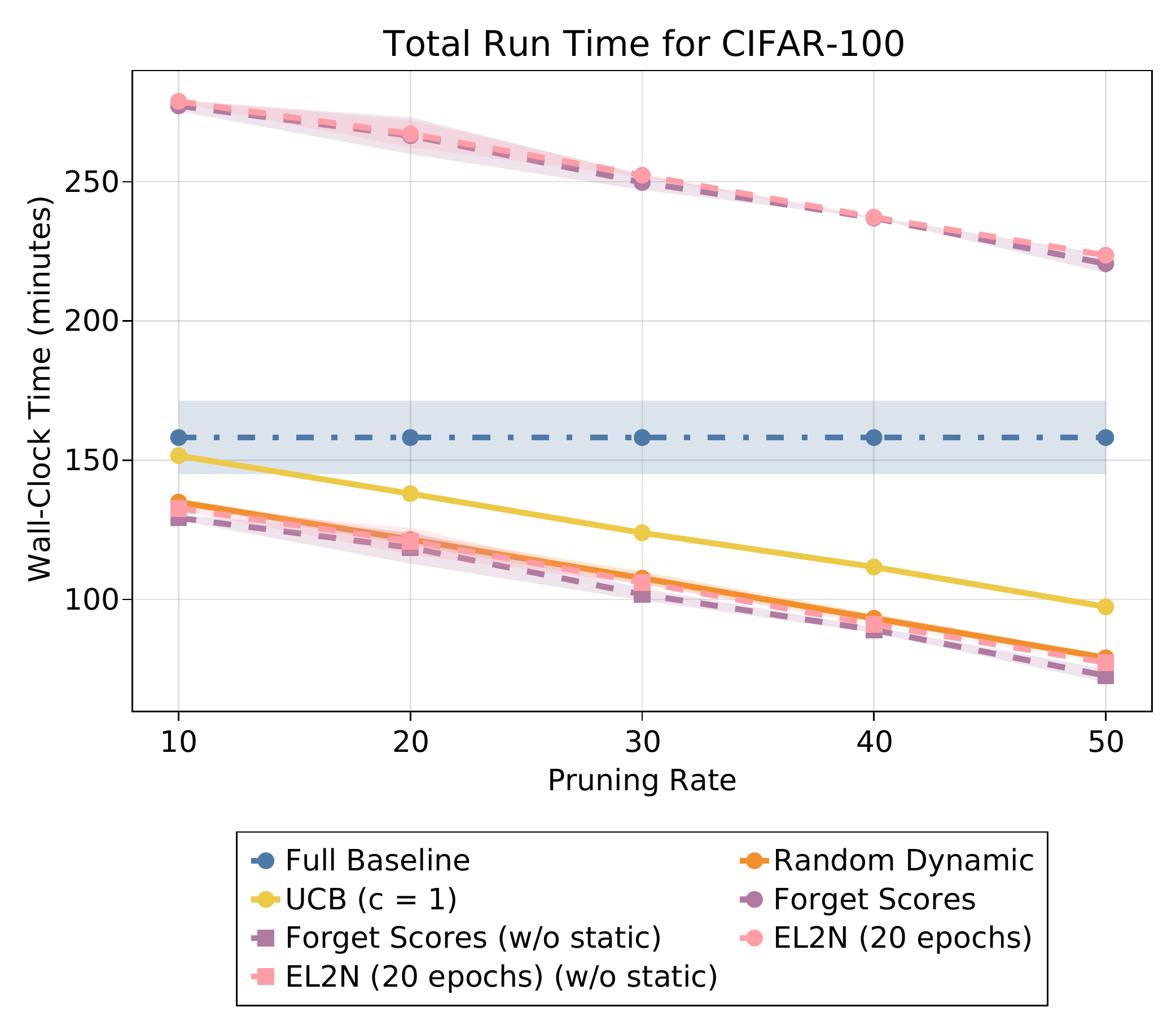}
    \caption{Dynamic data pruning on CIFAR-100 with pre-training scoring cost included and excluded for static methods.}
    \label{fig:static_cifar100_wc}
\end{figure}

Fig. \ref{fig:static_cifar10_wc} and Fig. \ref{fig:static_cifar100_wc} shows the wall-clock time of the static methods with and without the pre-training scoring cost. The static methods achieve roughly a similar run time to the random dynamic method, which is to be expected since there is no overhead at each checkpoint.

\section{Static approaches}
\begin{figure}
    \centering
    \includegraphics[width=0.45\textwidth]{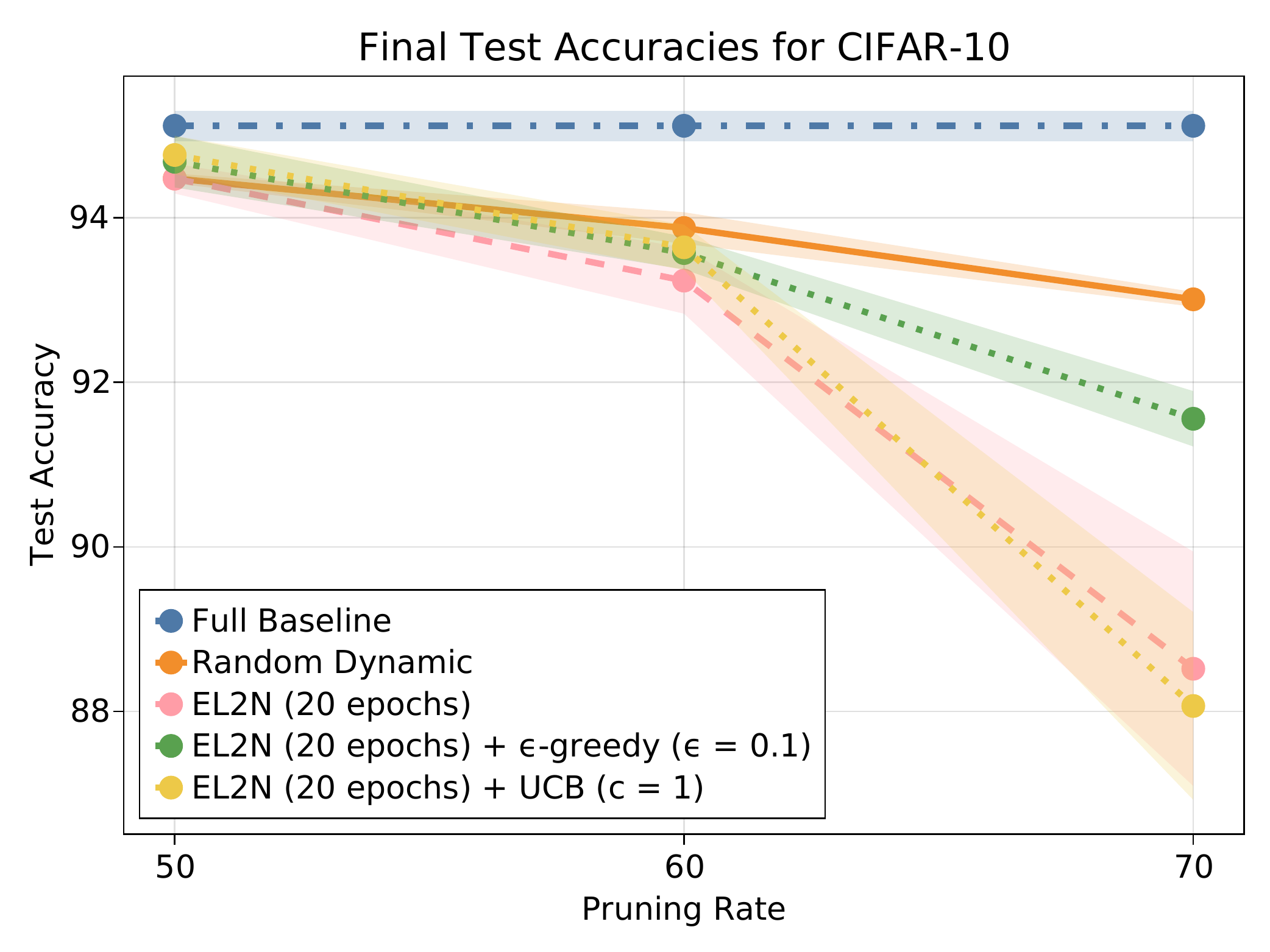}
    \caption{Apply our RL methods on top the static EL2N scores. ``EL2N + $\epsilon$-greedy'' uses the static EL2N scores at each checkpoint but also selects $\epsilon$ fraction of the samples randomly. ``EL2N + UCB'' applies the UCB algorithm statically to the EL2N scores from each model initialization trial.}
    \label{fig:static_methods}
\end{figure}

In our dynamic methods, we maintain running means and variances of the scores. In contrast, the static EL2N method takes an offline mean across many model initializations. A natural question is whether our methods can be applied in a static fashion for the same accuracy gains. The results in \cref{fig:static_methods} show that this is not the case.

When we statically apply the UCB algorithm to various model initialization trials in the EL2N method, we see no improvement in performance. This makes sense, because the variance in the UCB method is used to dynamically sample under-observed points. In a static variation, this feature of the algorithm is unused.

If we apply an $\epsilon$-greedy approach to the static EL2N policy (i.e. every checkpoint we select $1 - \epsilon$ fraction of the samples using the pre-computed EL2N scores, and we select $\epsilon$ fraction randomly), we see an increase in accuracy. Therefore, we see that even a small amount of dynamism improves static scores. The boost in accuracy still under-performs compared to the random dynamic method.

\section{Hyperparameter search}
\begin{figure}
    \centering
    \begin{subfigure}{0.45\textwidth}
        \includegraphics[width=\textwidth]{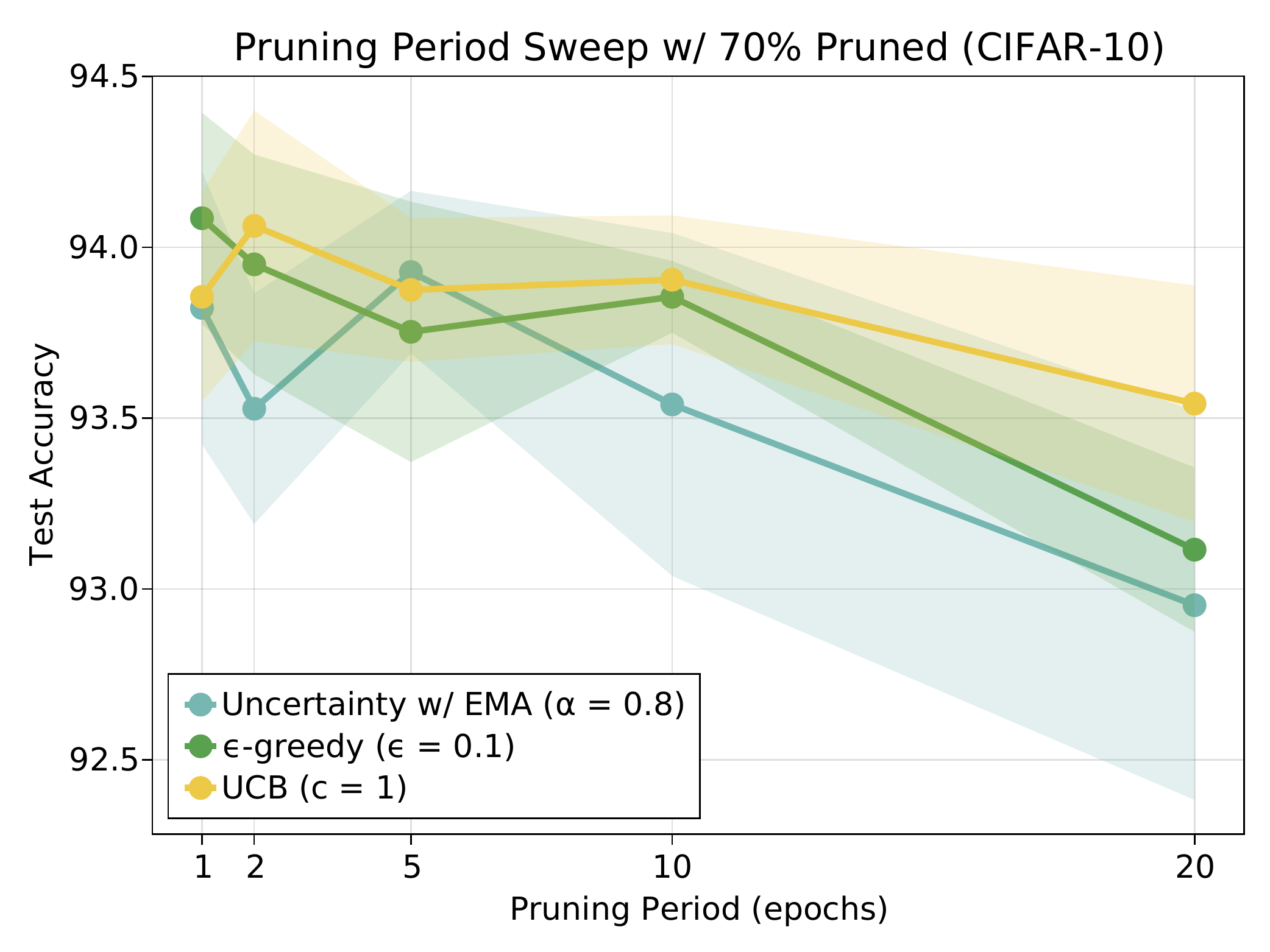}
        \caption{Sweeping the frequency for all methods.}
        \label{fig:cifar10-freq}
    \end{subfigure}
    \begin{subfigure}{0.45\textwidth}
        \includegraphics[width=\textwidth]{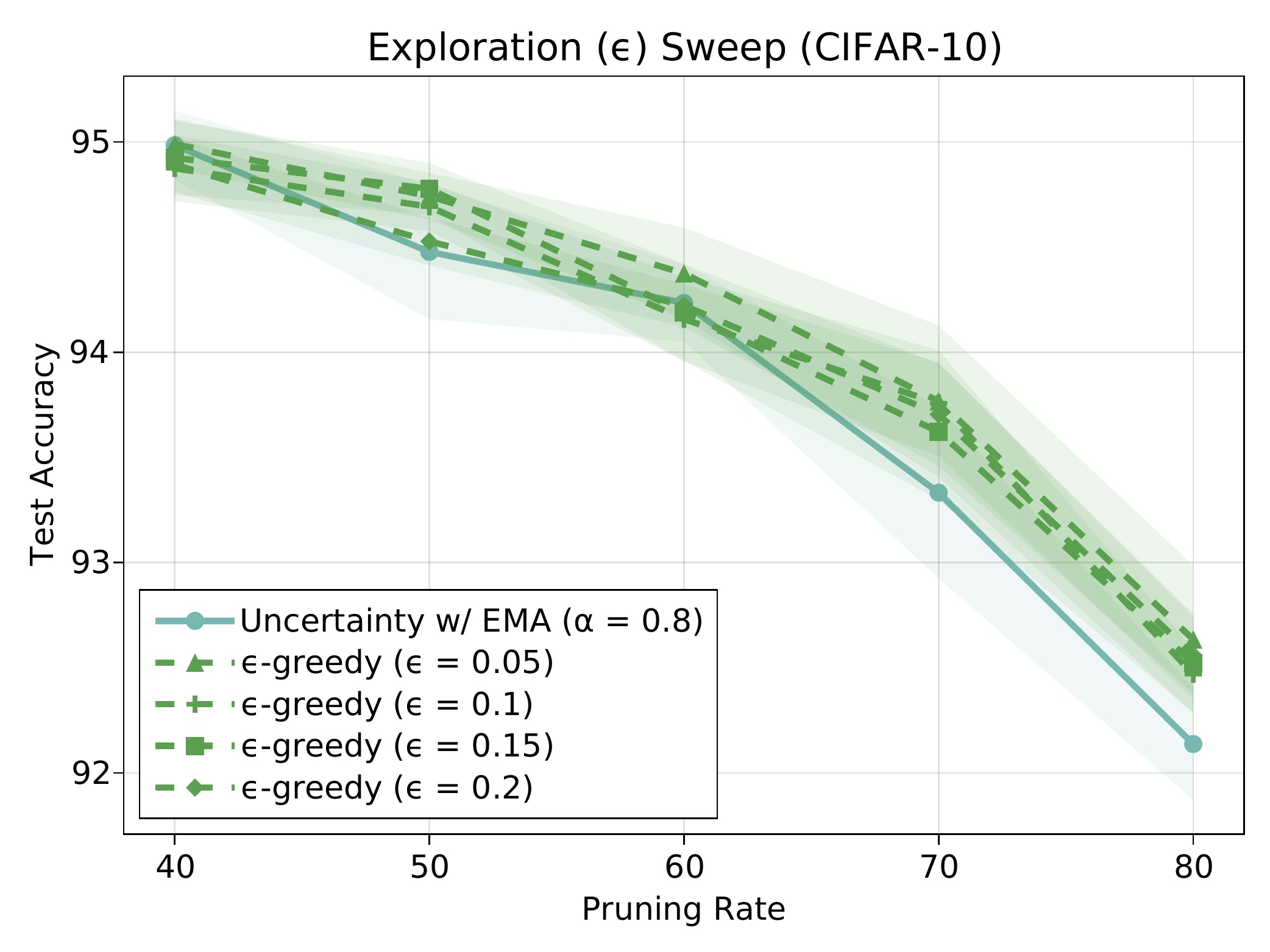}
        \caption{Sweeping the exploration rate ($\epsilon$) for the $\epsilon$-greedy method.}
        \label{fig:cifar10-eps}
    \end{subfigure}
    \begin{subfigure}{0.45\textwidth}
        \includegraphics[width=\textwidth]{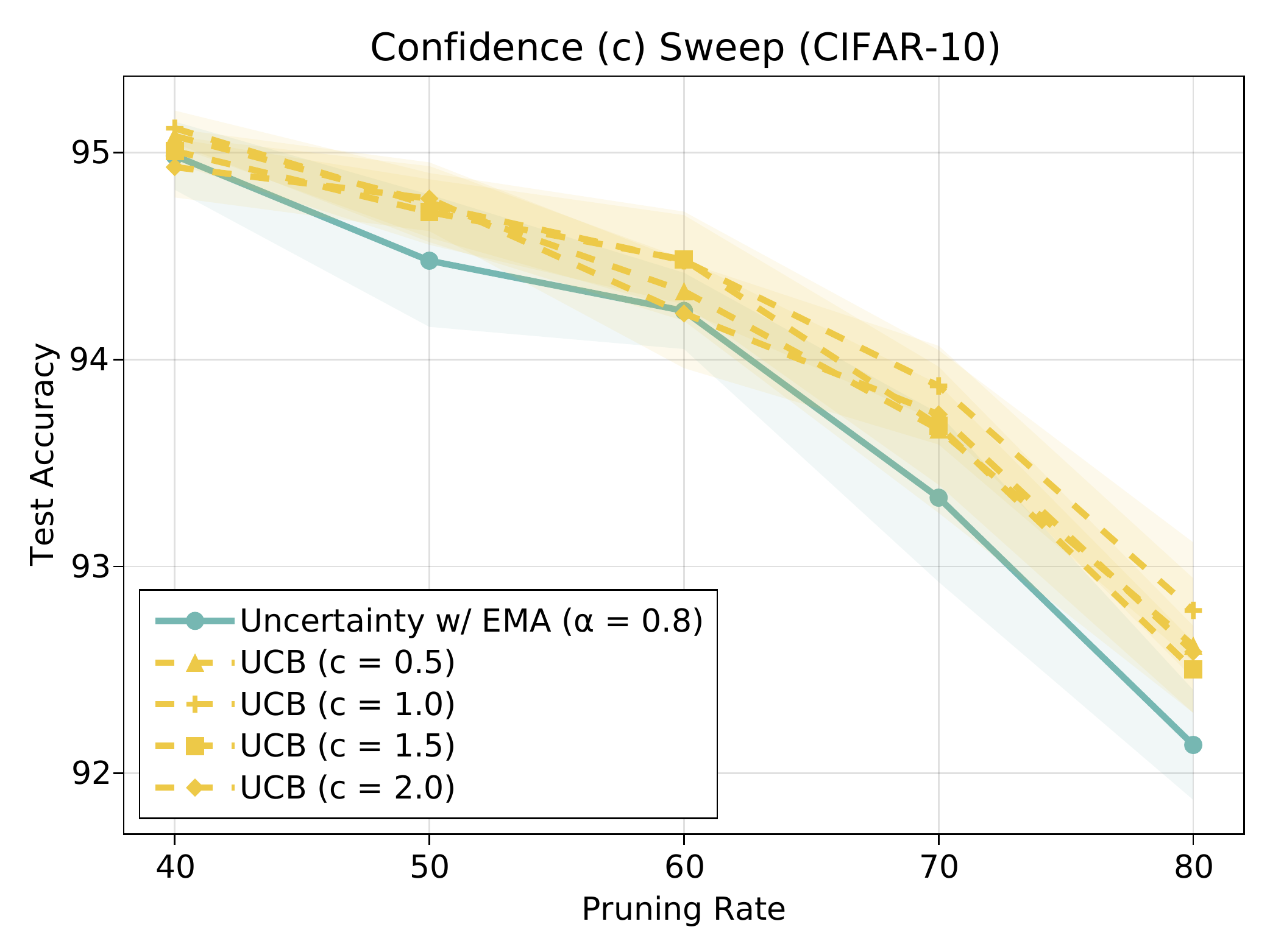}
        \caption{Sweeping the confidence ($c$) for the UCB method.}
        \label{fig:cifar10-confidence}
    \end{subfigure}
    \caption{The effect of sweeping various hyper-parameters while training on CIFAR-10.}
    \label{fig:cifar10-hyperparams}
\end{figure}

Fig. \ref{fig:cifar10-hyperparams} shows the results of sweeping the pruning period ($T_p$), the exploration rate ($\epsilon$; $\epsilon$-greedy only) and confidence ($c$; UCB only). The pruning period of 10 epochs provides the best performance vs. wall-clock time tradeoff for all methods. Varying the exploration rate does not impact the accuracy, suggesting that only a small amount of randomness is required. Varying the confidence suggests that incorporating variance into the scoring mechanism is sufficient to increasing the performance, agnostic of $c$'s value.

\section{Reproducing the results}
To run the code, change directories to the \texttt{dynamic\_data\_pruning\_code} folder. Follow the instructions in the \texttt{README.md} to re-create the training environment and run each experiment.

\end{document}